\documentclass[runningheads]{llncs}

\usepackage{eccv}

\usepackage{epsfig}
\usepackage{graphicx}
\usepackage{amsmath}
\usepackage{amssymb}
\usepackage{booktabs}
\usepackage{bbm}
\usepackage{bm}
\usepackage{multirow}
\usepackage{adjustbox}
\usepackage{pifont}
\usepackage{colortbl}
\usepackage[accsupp]{axessibility}  
\usepackage[title]{appendix}

\usepackage{eccvabbrv}
\usepackage{verbatim}

\usepackage{xcolor}
\definecolor{fit_curve_green}{RGB}{169,209,142}
\definecolor{supple_pink}{RGB}{245,151,155}
\definecolor{supple_blue}{RGB}{68,114,196}

\usepackage{graphicx}
\usepackage{booktabs}

\usepackage[accsupp]{axessibility}  

\usepackage{hyperref}

\usepackage{orcidlink}

\usepackage[normalem]{ulem}

\begin{document}

\title{HAC: Hash-grid Assisted Context for 3D Gaussian Splatting Compression}



\author{
Yihang Chen\inst{1, 2} \and
Qianyi Wu\inst{2} \and
Weiyao Lin\inst{1}\thanks{Corresponding author} \and \\
Mehrtash Harandi\inst{2} \and
Jianfei Cai\inst{2}
}

\authorrunning{Y.~Chen et al.}

\institute{Shanghai Jiao Tong University \and
Monash University \\
\email{\{yhchen.ee, wylin\}@sjtu.edu.cn, \\
\{qianyi.wu, mehrtash.harandi, jianfei.cai\}@monash.edu}
}

\maketitle

\begin{figure}
    \centering
    \includegraphics[width=0.95\linewidth]{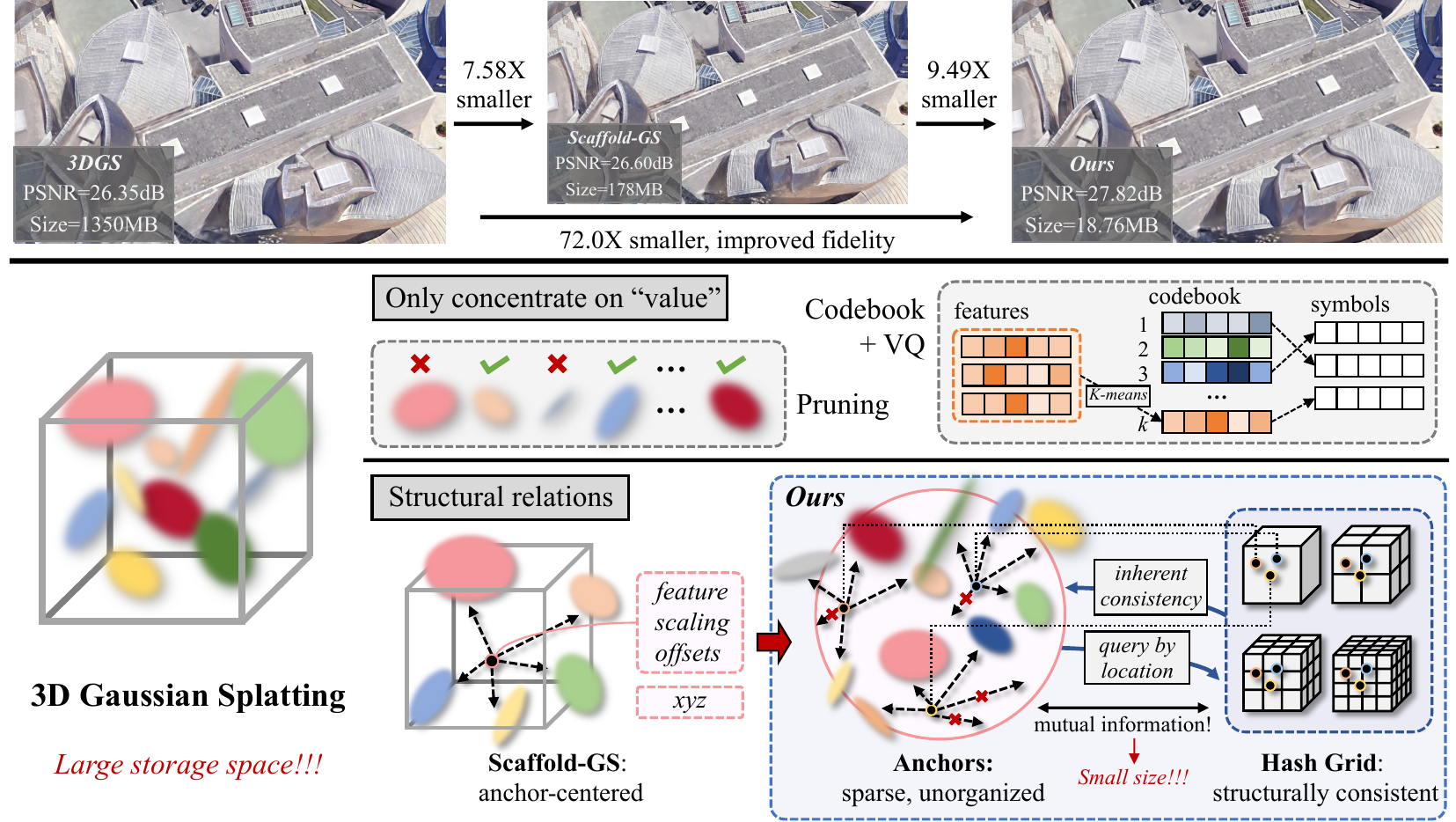}
    \caption{\textbf{Top}: A toy example where our method makes the size of the vanilla 3D Gaussian splitting (3DGS) model $72\times$ smaller (or $9.49\times$ smaller compared to the SoTA Scaffold-GS~\cite{scaffold}), with similar or better fidelity. \textbf{Bottom}: Most existing 3DGS compression methods 
    concentrate solely on parameter ``values'' using pruning or vector quantization to reduce size, ignoring the structure relations among Gaussians. Scaffold-GS~\cite{scaffold} introduces anchors to cluster and neural-predict the associated Gaussians while treating each anchor point independently. Our core idea is to further exploit the inherent consistencies of anchors via a structured hash grid for a more compact 3DGS representation.     
    }
    \label{fig:teaser}
\end{figure}

\begin{abstract}
  
  3D Gaussian Splatting (3DGS) has emerged as a promising framework for novel view synthesis, boasting rapid rendering speed with high fidelity. However, the substantial Gaussians and their associated attributes necessitate effective compression techniques. Nevertheless, the sparse and unorganized nature of the point cloud of Gaussians (or anchors in our paper) presents challenges for compression. To address this, we make use of the relations between the unorganized anchors and the structured hash grid, leveraging their mutual information for context modeling, and propose a Hash-grid Assisted Context (HAC) framework for highly compact 3DGS representation.
  Our approach introduces a binary hash grid to establish continuous spatial consistencies, allowing us to unveil the inherent spatial relations of anchors through a carefully designed context model. To facilitate entropy coding, we utilize Gaussian distributions to accurately estimate the probability of each quantized attribute, where an adaptive quantization module is proposed to enable high-precision quantization of these attributes for improved fidelity restoration. Additionally, we incorporate an adaptive masking strategy to eliminate invalid Gaussians and anchors. 
  Importantly, our work is the pioneer to explore context-based compression for 3DGS representation, resulting in a remarkable size reduction of over $75\times$ compared to vanilla 3DGS, while simultaneously improving fidelity, and achieving over $11\times$ size reduction over SoTA 3DGS compression approach Scaffold-GS. Our code is available \href{https://github.com/YihangChen-ee/HAC}{\textcolor{red}{here}}.

  \keywords{3D Gaussian Splatting \and Compression \and Context Models}
\end{abstract}

\section{Introduction}
\label{sec:intro}


Over the past few years, significant advancements have been made in 3D scene representations for novel view synthesis.
Neural Radiance Field (NeRF)~\cite{NeRF} proposes rendering colors by accumulating RGB values along sampling rays using an implicit Multilayer Perceptron (MLP), aiming at reconstructing photo-realistic images. However, the extensive sampling of ray points has been a bottleneck, affecting both the speed of training and rendering. Recent advances of NeRF~\cite{INGP, TensoRF, K-planes} introduce feature grids to enhance the rendering process, facilitating faster rendering speeds by reducing the MLP size. Despite the improvement, these approaches still suffer from relatively slow rendering speeds due to frequent ray point sampling.

In this context, very recently, a new
paradigm of 3D representation, 3D Gaussian Splatting (3DGS)~\cite{3DGS}, emerged. 
%
3DGS introduces learnable Gaussians to directly represent 3D space explicitly. These Gaussians, initialized from Structure-from-Motion (SfM)~\cite{SfM}
and endowed with learnable shape and appearance parameters, can be directly splatted onto 2D planes for rapid and differentiable rendering within imperceptible intervals using tile-based rasterization~\cite{raster}. As such, the time-consuming volume rendering used in NeRF can be completely removed. The advantages of rapid differentiable rendering with high photo-realistic fidelity have stimulated the fast and widespread adoption of 3DGS in the field.

However, 3DGS is not the ultimate solution. One major drawback is that it requires a considerable number of 3D Gaussians to well represent a large-scale scene (\eg, at the scale of millions of Gaussians for city-scale scenes) and needs a large storage space (\eg, a few  GigaBytes (GB)) to store the associated Gaussian attributes for each scene~\cite{BungeeNeRF}. This motivates us to investigate effective compression techniques for 3DGS.


Due to their sparse and unorganized nature, compressing 3D Gaussians is challenging and difficult~\cite{chen2024survey, fei2024survey}.
Therefore, most existing 3DGS compression approaches focus solely on parameter ``values'' but overlook their structural relations. For example, as illustrated in~\cref{fig:teaser} middle, parameter pruning can be used to mask out the Gaussians whose parameter values are below a certain threshold \cite{Joo,zhiwen}.
%
Another straightforward technique is to apply vector quantization to cluster parameters with similar ``values''. 
Such an approach enables the direct compression of parameters by only retaining more representative ones while maintaining reconstruction fidelity~\cite{Joo, Simon, KLNavaneet, zhiwen}. Nevertheless, solely concentrating on ``values'' fails to 
eliminate structural redundancies, which are pivotal for compact representations. To exploit such spatial relations of Gaussians, 
Scaffold-GS~\cite{scaffold} introduces anchors to cluster related nearby 3D Gaussians and neural-predict their attributes from the anchors' attributes, resulting in significant storage savings. Despite the improvement, Scaffold-GS still treats each anchor independently, and there are still substantial anchors that are sparse, unorganized, and hard to compress, due to their point-cloud nature.
%

To further push the boundary of 3DGS compression, we draw inspiration from the NeRF series~\cite{NeRF}, contemplating the idea of representing 3D space using well-organized feature grids~\cite{INGP, TensoRF}. We pose the question: \emph{Is there inherent relations between the attributes of unorganized anchors in Scaffold-GS and the structured feature grids?} Our answer is affirmative since we observe large mutual information between anchor attributes and the hash grid features. Based on this observation, we 
%
propose a Hash-grid Assisted Context (HAC) framework, where our core idea is to jointly learn structured compact hash grid (binarized for each hash parameter) and use it for context modeling of anchor attributes. 
%
Specifically, with Scaffold-GS~\cite{scaffold} as our base model, for each anchor, we query the hash grid by the anchor location to obtain an interpolated hash feature, which is then used to 
predict the value distributions of anchor attributes, facilitating entropy coding such as Arithmetic Coding (AE)~\cite{AE} for a highly compact representation of the model. 
Note that we employ Scaffold-GS as our base model as its anchor-centered design provides a good foundation to establish relations with these interpolated hash features.
Furthermore, we introduce an Adaptive Quantization Module (AQM), which dynamically adjusts different quantization step sizes for different anchor attributes for retaining of their original information. Learnable masks are also employed to mask out invalid Gaussians and anchors, further enhancing the compression ratio.
Our main contributions can be summarized as follows:
\begin{enumerate}
    \item To our knowledge, we are the first to model contexts for 3DGS compression, \ie, using a structured hash grid to exploit the inherent consistencies among unorganized 3D Gaussians (or anchors in Scaffold-GS).

    \item To facilitate efficient entropy encoding of anchor attributes, we propose to use the interpolated hash feature to neural-predict the value distribution of anchor attributes as well as neural-predicting quantization step refinement with AQM. We also employ learnable masks to prune out ineffective Gaussians and anchors. 
    

    \item Extensive experiments on five datasets demonstrate the effectiveness of our HAC framework and each technical component. We achieve a compression ratio of $11\times$ over our base model Scaffold-GS and $75\times$ over the vanilla 3DGS model when averaged over all datasets, while with comparable or even improved fidelity.
    
\end{enumerate}

\section{Related Work}
\noindent\textbf{Neural Radiance Field and its compression.} The emergence of Neural Radiance Field (NeRF)\cite{NeRF} has significantly advanced novel view synthesis by employing a single learnable implicit MLP to generate arbitrary views of 3D scenes through $\alpha$-composed accumulation of RGB values along a ray. However, the dense querying of sampling points and the utilization of a large MLP hinder real-time rendering. To address this problem, subsequent approaches such as Instant-NGP\cite{INGP}, TensoRF~\cite{TensoRF}, K-planes~\cite{K-planes}, and DVGO~\cite{DVGO} adopt explicit grid-based representations to facilitate faster training and rendering by reducing the size of the MLP, which however comes at the cost of increased storage space.

To mitigate the storage increase, compression techniques focusing on reducing the size of explicit representations have been developed, which can be categorized into either ``value''-based or structural-relation-based approaches. The former category includes pruning~\cite{VQRF, Re:NeRF}, codebooks~\cite{VQRF, CompactNeRF}, and methods like quantization or entropy constraint employed in BiRF~\cite{BiRF} and SHACIRA~\cite{SHACIRA}. On the other hand, the latter category explores structural relations via wavelet decomposition~\cite{MaskDWT}, rank-residual decomposition~\cite{CCNeRF}, or spatial prediction~\cite{SPC-NeRF} to eliminate spatial redundancy, thanks to the well-structured characteristics of these feature grids. CNC~\cite{cnc2024} provides a solid proof of concept by sufficiently utilizing such structural information, achieving remarkable RD performance gain.

\noindent\textbf{3D Gaussian Splatting and its compression.} 3DGS~\cite{3DGS} has innovatively addressed the challenge of slow training and rendering in NeRF while maintaining high-fidelity quality by representing 3D scenes with 3D Gaussians endowed with learnable shape and appearance attributes. By adopting differentiable splatting and tile-based rasterization~\cite{raster}, 3D Gaussians are optimized during training to best fit their local 3D regions. Despite its advantages, the substantial Gaussians and their associated attributes necessitate effective compression techniques.

Unlike NeRF-based feature grids, 3D Gaussians in 3DGS are sparse and unorganized, presenting significant challenges for establishing structural relations. Consequently, compression approaches have primarily focused solely on the ``value'' of model parameters, employing techniques such as pruning~\cite{Joo, zhiwen}, codebooks~\cite{Joo, Simon, zhiwen, KLNavaneet}, and entropy constraints~\cite{Sharath}. 
To our knowledge, Scaffold-GS~\cite{scaffold} and Morgenstern~\etal~\cite{Wieland} have explored the relations of Gaussians. In~\cite{scaffold}, authors introduce anchor-centered features to achieve reduced parameter numbers, while in~\cite{Wieland} dimension collapsing is considered to compress Gaussians in an ordered 2D space. However, their investigation of spatial redundancy remains insufficient.

In this paper, we emphasize leveraging such structural relations for compression is crucial. For instance, approaches in image compression~\cite{cheng2020learned, he2021checkerboard, he2022elic} and video compression~\cite{DCVC, DCVC-HEM, DCVC-TCM} have demonstrated the effectiveness of eliminating structural redundancy by excavating spatial and temporal relations, thanks to their well-organized data structure. 
Motivated by this, with Scaffold-GS as our base model, we introduce a well-structured hash grid as context to model the inherent consistencies of the sparse and unorganized anchors, 
%
achieving much more compact 3DGS representation.

\section{Methods}
\begin{figure}[t]
    \centering
    \includegraphics[width=1.0\linewidth]{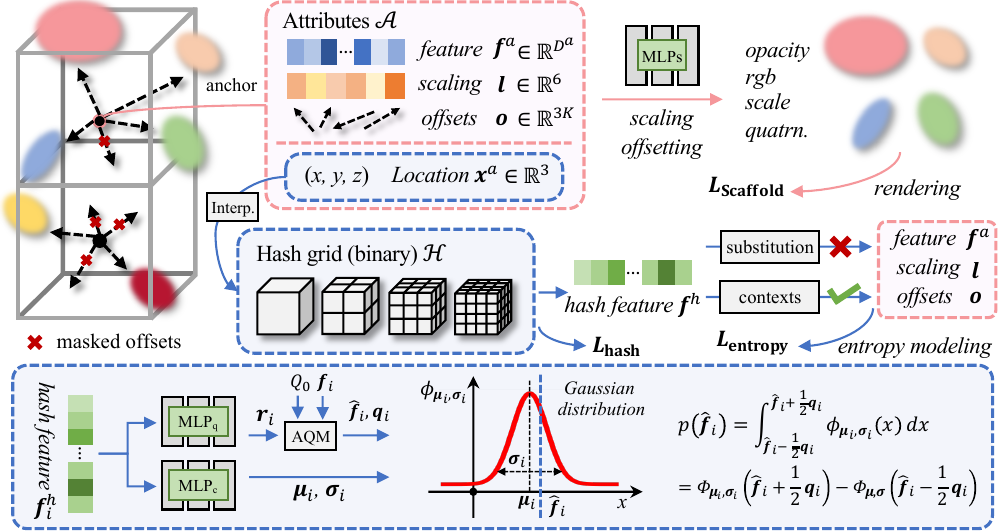}
    \caption{Overview of our HAC framework. It is based on Scaffold-GS~\cite{scaffold} (\textbf{top}), which introduces anchors with their attributes 
    to neural-predict 3D Gaussian attributes. 
    \textbf{Middle}: Our HAC framework jointly learns structured compact hash grid (binarized for each parameter) that can be queried at any anchor location to obtain the interpolated hash feature $\bm{f}^h$. Instead of direct substitution, $\bm{f}^h$ is used as context to predict the value distributions of anchor attributes, which is essential for the subsequent entropy coding.  
    \textbf{Bottom}: Our proposed context models take $\bm{f}^h$ as input and outputs ${\bm{r}}$ for the AQM (quantize anchor attribute values into a finite set) and the parameters ($\bm{\mu}$ and $\bm{\sigma}$) to model the value distributions of anchor attributes. 
        }
    \label{fig:main_method}
\end{figure}

In~\cref{fig:main_method}, we conceptualize our HAC framework. In particular, HAC is based on the baseline Scaffold-GS~\cite{scaffold} (\cref{fig:main_method}~{top}), which introduces anchors with their attributes $\mathcal{A}$ (feature, scaling and offsets) 
to cluster and neural-predict 3D Gaussian attributes (opacity, RGB, scale, and quaternion). At the core of our HAC, we propose to jointly learn structured compact hash grid (binarized for each parameter) that can be queried at any anchor location to obtain the interpolated hash feature $\bm{f}^h$ (\cref{fig:main_method}~{middle}). Instead of directly substituting anchor feature, $\bm{f}^h$ is used as context to predict the value distributions of anchor attributes, which is essential for the subsequent entropy coding, \ie, Arithmetic Coding (AE). Our context model (\cref{fig:main_method} bottom) is a simple MLP that takes $\bm{f}^h$ as input and outputs ${\bm{r}}$ for the adaptive quantization module (AQM) (quantize anchor attribute values into a finite set) and the Gaussian parameters ($\bm{\mu}$ and $\bm{\sigma}$) for modeling the value distributions of anchor attributes, from which we can compute the probability of each quantized attribute value for AE. Note that we draw two MLPs (${\rm MLP_q}$ and ${\rm MLP_c}$) in~\cref{fig:main_method} for easy explanation but they actually share the same MLP layers with outputs at different dimensions. 
Besides, an adaptive offset masking module (\cref{fig:main_method} top-left) is adopted to prune redundant Gaussians and anchors.
In the following, we first introduce the background and then delve into the detailed technical components of our HAC.


\subsection{Preliminaries}

\noindent\textbf{3D Gaussian Splatting (3DGS)}~\cite{3DGS} represents a 3D scene using numerous Gaussians and renders viewpoints through a differentiable splatting and tile-based rasterization. Each Gaussian is initialized from SfM and defined by a 3D covariance matrix $\bm{\Sigma}\in\mathbb{R}^{3\times3}$ and location (mean) $\bm{\mu}\in\mathbb{R}^{3}$,

\begin{equation}
    G(\bm{x}) = \exp{\left(-\frac{1}{2}(\bm{x}-\bm{\mu})^\top\bm{\Sigma}^{-1}(\bm{x}-\bm{\mu})\right)}\;,
\end{equation}
where $\bm{x}\in\mathbb{R}^{3}$ is a random 3D point, and $\bm{\Sigma}$ is defined by a diagonal matrix $\bm{S}\in\mathbb{R}^{3\times3}$ representing scaling
and rotation matrix $\bm{R}\in\mathbb{R}^{3\times3}$ to guarantee its positive semi-definite characteristics, such that $\bm{\Sigma}=\bm{R}\bm{S}\bm{S}^\top\bm{R}^\top$.
To render an image from a random viewpoint, 3D Gaussians are first splatted to 2D, and render the pixel value $\bm{C}\in\mathbb{R}^{3}$ using $\alpha$-composed blending,

\begin{equation}
    \bm{C} = \sum_{i\in I} {\bm{c}_i\alpha_i\prod_{j=1}^{i-1}\left(1-\alpha_j\right)}
\end{equation}
where $\alpha\in\mathbb{R}^1$
measures the opacity of each Gaussian after 2D projection, $\bm{c}\in\mathbb{R}^{3}$ is view-dependent color modeled by Spherical Harmonic (SH) coefficients, and $I$ is the number of sorted Gaussians contributing to the rendering.

\noindent\textbf{Scaffold-GS}~\cite{scaffold} adheres to the framework of 3DGS and introduces a more storage-friendly and fidelity-satisfying anchor-based approach. It utilizes anchors to cluster Gaussians and deduce their attributes from the attributes of attached anchors through MLPs, rather than directly storing them. Specifically, each anchor consists of a location $\bm{x}^a\in\mathbb{R}^3$ and anchor attributes $\mathcal{A}=\{\bm{f}^a\in\mathbb{R}^{D^a}, \bm{l}\in\mathbb{R}^6, \bm{o}\in\mathbb{R}^{3K}\}$, where each component represents anchor feature, scaling and offsets, respectively. During rendering, $\bm{f}^a$ is inputted into MLPs to generate attributes for Gaussians, whose locations are determined by adding $\bm{x}^a$ and $\bm{o}$, where $\bm{l}$ is utilized to regularize both locations and shapes of the Gaussians. While Scaffold-GS has demonstrated effectiveness via this anchor-centered design, we contend there is still significant redundancy among inherent consistencies of anchors that we can fully exploit for a more compact 3DGS representation.

\subsection{Bridging Anchors and Hash Grid}
\begin{figure}[t]
    \begin{minipage}[c]{0.45\linewidth}
        \centering
        \includegraphics[width=5.0cm]{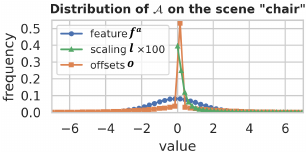}
    \end{minipage}%
    \begin{minipage}[c]{0.55\linewidth}
        \vfill
        \centering
        {\scriptsize
        \begin{tabular}{cccc}
        \toprule
        {\bf Synthetic-NeRF~\cite{NeRF}} & PSNR$\uparrow$ & SSIM$\uparrow$ & LPIPS$\downarrow$\\
        \midrule
        3DGS~\cite{3DGS} & 33.80 & 0.970 & 0.031\\
        Scaffold-GS~\cite{scaffold} & 33.41 & 0.966 & 0.035\\
        Substituting $\bm{f}^a$ with $\bm{f}^h$ & 32.85 & 0.963 & 0.041\\
        \bottomrule
        \end{tabular}
        }
        \vfill
    \end{minipage}
    \caption{\textbf{Left chart}: Statistical analysis of the value distributions of $\mathcal{A}$ on the scene ``chair'' of the Synthetic-NeRF dataset~\cite{NeRF}. All three components $\{\bm{f}^a$, $\bm{l}, \bm{o}\}$ exhibit statistical Gaussian distributions. Note that the values of $\bm{l}$ are scaled by a factor of 100 for better visualization. \textbf{Right table}: Experimental results of directly substituting anchor feature $\bm{f}^a$ with hash feature $\bm{f}^h$ on this dataset.}
    \label{fig:hash_gaussian_mask}
\end{figure}

We begin the analysis by intuitively considering neighboring Gaussians share similar parameters inferred from anchor attributes. This initial perception leads us to assume anchor attributes are also consistent in space. Our main idea is to leverage the well-structured hash grid to unveil the inherent spatial consistencies of the unorganized anchors. Please also refer to experiments in~\cref{fig:bit_allocation} to observe this consistency.
To verify mutual information between the hash grid and anchors, we first explore substituting 
anchor features $\bm{f}^a$ with hash features $\bm{f}^h$ that are acquired by interpolation using the anchor location $\bm{x}^a$ on the hash grid $\mathcal{H}$,
defined as $\bm{f}^h:= {\rm Interp}(\bm{x}^a, \mathcal{H})$. Here, $\mathcal{H}=\{\bm{\theta}_i^l\in\mathbb{R}^{D^h}|i=1,\dots,T^l|l=1,\dots,L\}$ represents the hash gird, where $D^h$ is the dimension of vector $\bm{\theta}_i^l$, $T^l$ is the table size of the grid for level $l$, and $L$ is the number of levels.
We conduct a preliminary experiment on the Synthetic-NeRF dataset~\cite{NeRF} to assess its performance, as shown in the right panel of~\cref{fig:hash_gaussian_mask}. Direct substitution using hash features appears to yield inferior fidelity and introduces drawbacks such as unstable training (due to its impact on anchor spawning processes) and decreased testing FPS (owing to the extra interpolation operation). These results may further degrade if $\bm{l}$ and $\bm{o}$ are also substituted for a more compact model. Nonetheless, we find the fidelity degradation remains moderate, suggesting the existence of rich mutual information between $\bm{f}^h$ and $\bm{f}^a$. This prompts us to ask: \emph{Can we exploit such mutual relation and use the compact hash features to model the context of anchor attributes $\mathcal{A}$?}
This leads to the context modeling as a conditional probability:
\begin{equation}
\label{eq:prior_probability}
    p(\mathcal{A}, \bm{x}^a, \mathcal{H}) = p(\mathcal{A}|\bm{x}^a, \mathcal{H})\times p(\bm{x}^a, \mathcal{H}) \sim p(\mathcal{A}|\bm{f}^h)\times p(\mathcal{H})
\end{equation}
where $\bm{x}^a$ is omitted in the last term as we assume the independence of $\bm{x}^a$ and $\mathcal{H}$ (it can be anywhere), making $p(\mathcal{H}|\bm{x}^a) \sim p(\mathcal{H})$, and do not employ entropy constraints to $\bm{x}^a$.
According to information theory~\cite{Elements_of_information_theory}, a higher probability corresponds to lower uncertainty (entropy) and fewer bits consumption. Thus, the large mutual information between $\mathcal{A}$ and $\bm{f}^h$ ensures a large $p(\mathcal{A}|\bm{f}^h)$. Our goal is to devise a solution to effectively leverage this relationship. Furthermore, $p(\mathcal{H})$ signifies that the size of the hash grid itself should also be compressed, which can be done by adopting the existing solution for Instant-NGP compression~\cite{cnc2024}. 

We underscore the significance of this conditional probability based approach since it ensures both rendering speed and fidelity upper-bound unaffected as it only utilizes hash features to estimate the entropy of anchor attributes for entropy coding but does not modify the original Scaffold-GS structure. In the following subsections, we delve into the technical details of our context models.

\subsection{HAC: Hash-Grid Assisted Context Framework}
The principle objective of HAC is to minimize the entropy of anchor attributes $\mathcal{A}$ with the assistance of hash feature $\bm{f}^h$ (\ie, maximize $p(\mathcal{A}|\bm{f}^h)$), facilitating bit reduction when encoding anchor attributes using entropy coding like AE\cite{AE}.
As shown in~\cref{fig:main_method}, anchor locations $\bm{x}^a$ are firstly inputted into the hash grid for interpolation, the obtained $\bm{f}^h$ are then employed as context for $\mathcal{A}$. 


\noindent\textbf{Adaptive Quantization Module}. 
To facilitate entropy coding, values of $\mathcal{A}$ must be quantized to a finite set. Our empirical studies reveal that binarization, as that in BiRF~\cite{BiRF}, is unsuitable for $\mathcal{A}$ as it fails to preserve sufficient information. Thus, we opt for rounding them to maintain their comprehensive features. To ensure backpropagation, we utilize the ``adding noise'' operation during training and ``rounding'' during testing, as described in \cite{balle2018variational}.

Nevertheless, the conventional rounding is essentially a quantization with a step size of ``1'', which is inappropriate for the scaling $\bm{l}$ and the offset $\bm{o}$, since they are usually decimal values. To address this, we further introduce an  Adaptive Quantization Module (AQM), which adaptively determines quantization steps. 
In particular, for the $i$th anchor $\bm{x}^a_i$, we denote $\bm{f}_i$ as any of its $\mathcal{A}_i$'s components: $\bm{f}_i\in{\{\bm{f}^a_i, \bm{l}_i, \bm{o}_i\}}\in\mathbb{R}^D$, where $D\in\{D^a, 6, 3K\}$ is its respective dimension.
The quantization can be written as,
\begin{equation}
\label{eq:quantize_2}
\begin{aligned}
    \hat{\bm{f}_i} &= \bm{f}_i + \mathcal{U}\left(-\frac{1}{2}, \frac{1}{2}\right)\times \bm{q}_i , \qquad\;\;\; \text{for training} \\
    &= {\rm Round}(\bm{f}_i / \bm{q}_i)\times \bm{q}_i , \qquad \qquad \, \text{for testing}\\
\end{aligned}
\end{equation}
where
\begin{equation} \label{eq:qi}
\begin{aligned}
    \bm{q}_i &= Q_0\times\left(1+{\rm Tanh}\left({\bm{r}}_i\right)\right) \\
    {\bm{r}}_i &= {\rm MLP_q}\left(\bm{f}^h_i\right).
\end{aligned}
\end{equation}
We use a simple MLP-based context model ${\rm MLP_q}$ 
to predict from hash feature $\bm{f}^h_i$ a refinement ${\bm{r}}_i\in\mathbb{R}^1$, which is used to adjust the predefined quantization step size $Q_0$. Note that $Q_0$ varies for $\bm{f}^a$, $\bm{l}$, and $\bm{o}$. \cref{eq:qi} essentially restricts the quantization step size $\bm{q}_i\in\mathbb{R}^1$ to be chosen within $\left(0, 2Q_0\right)$, enabling $\hat{\bm{f}}_i$ to closely resemble the original characteristics of $\bm{f}_i$, 
maintaining a high fidelity. 


\noindent\textbf{Gaussian Distribution Modeling}. To measure the bit consumption of $\hat{\bm{f}}_i$ during training, its probability needs to be estimated in a differentiable manner. 
As shown in~\cref{fig:hash_gaussian_mask} left, all three components of anchor attributes $\mathcal{A}$ exhibit statistical tendencies of Gaussian distributions, where $\bm{l}$ displays a single-sided pattern due to Sigmoid activation\footnote[1]{We define $\bm{l}$ as the one \textit{after} Sigmoid activation, which is slightly different from \cite{scaffold}.}. 
This observation establishes a lower bound for probability prediction when all $\hat{\bm{f}}_i$s in $\mathcal{A}$ are estimated using the respective $\mu$ and $\sigma$ of the statistical Gaussian Distribution of $\bm{f}^a$, $\bm{l}$ and $\bm{o}$.
Nevertheless, employing a single set of $\mu$ and $\sigma$ for all attributes may lack accuracy. Therefore, we assume anchor attributes $\mathcal{A}$'s values independent, and construct their respective Gaussian distributions, where their individual $\bm{\mu}$ and $\bm{\sigma}$ are estimated by a simple MLP-based context model ${\rm MLP_c}$ from $\bm{f}^h$. 
%
Specifically, for the $i$th anchor and its quantized anchor attribute vector $\hat{\bm{f}}_i$, with the estimated $\bm{\mu}_i\in\mathbb{R}^D$ and $\bm{\sigma}_i\in\mathbb{R}^D$, we can compute the probability of $\hat{\bm{f}}_i$ as, 
%
%
%
\begin{equation}
\begin{aligned}
    p(\hat{\bm{f}}_i) &= \int_{\hat{\bm{f}}_i-\frac{1}{2}\bm{q}_i}^{\hat{\bm{f}}_i+\frac{1}{2}\bm{q}_i}\phi_{\bm{\mu}_i, \bm{\sigma}_i}\left(x\right)\,dx \\
    &= \Phi_{\bm{\mu}_i, \bm{\sigma}_i}\left(\hat{\bm{f}}_i+\frac{1}{2}\bm{q}_i\right) - \Phi_{\bm{\mu}_i, \bm{\sigma}_i}\left(\hat{\bm{f}}_i-\frac{1}{2}\bm{q}_i\right) \\
    \bm{\mu}_i, \bm{\sigma}_i &= {\rm MLP_c}\left(\bm{f}^h_i\right).
\end{aligned}
\end{equation}
%
%
where $\phi$ and $\Phi$ represent the probability density function and the cumulative distribution function, respectively. Consequently, we define an entropy loss  as the summation of bit consumption over all $\bm{\hat{f}}_i$s:
\begin{equation}
    L_{\text{entropy}} = \sum_{\bm{f}\in\{\bm{f}^a, \bm{l}, \bm{o}\}} \sum_{i=1}^N \sum_{j=1}^D \left(-\log_2 p(\hat{{f}}_{i,j})\right)
\end{equation}
$N$ is the number of anchors and $\hat{{f}}_{i,j}$ is $\bm{\hat{f}}_i$'s $j$-th dimension value. Minimizing the entropy loss encourages a high probability estimation for $p(\hat{\bm{f}}_i)$, which in turn encourages accurate $\bm{\mu}_i$ and small $\bm{\sigma}_i$, guiding the learning of the context model.


%

\noindent\textbf{Adaptive Offset Masking}. 
From~\cref{fig:hash_gaussian_mask} left, we can also see that $\bm{o}$ exhibits an impulse at zero, suggesting the occurrence of substantial unnecessary Gaussians. 
%
Thus, we employ the technique introduced by Lee~\etal~\cite{Joo} to prune invalid $\bm{o}$ by utilizing straight-through~\cite{STE} estimated binary masks. Specifically, we apply the same marking loss $L_m$ in~\cite{Joo}  to encourage masking as many Gaussians as possible. This process effectively masks out invalid offsets and saves storage space directly. Additionally, we implement anchor pruning: if all the attached $\bm{o}$ are pruned on an anchor, then this anchor no longer contributes to rendering and should be pruned entirely (including its $\bm{x}^a$ and $\mathcal{A}$). 

\noindent\textbf{Hash Grid Compression}.
As shown in~\cref{eq:prior_probability}, the size of the hash grid $\mathcal{H}$ also significantly influences the final storage size. To this end, we binarize the hash table to $\{-1, +1\}$ using straight-through estimation (STE)~\cite{BiRF} and calculate the occurrence frequency $h_f$~\cite{cnc2024} of the symbol ``$+1$'' to estimate its bit consumption:
\begin{equation}
    L_{\text{hash}} = M_+\times\left(-\log_2(h_f)\right) + M_-\times\left(-\log_2(1-h_f)\right)
\end{equation}
where $M_+$ and $M_-$ are total numbers of ``$+1$'' and ``$-1$'' in the hash grid. 

\subsection{Training and Coding Process}
During training, we incorporate both the rendering fidelity loss and the entropy loss to ensure the model improves rendering quality while controlling total bitrate consumption in a differentiable manner. Our overall loss is
\begin{equation} \label{eq:loss}
    Loss = L_{\text{Scaffold}} + \lambda_e\frac{1}{N(D^a+6+3K)}(L_{\text{entropy}} + L_{\text{hash}}) + \lambda_m L_m .
\end{equation}
Here, $L_{\text{Scaffold}}$ represents the rendering loss as defined in~\cite{scaffold}, which includes two fidelity penalty loss terms and one regularization term for the scaling $\bm{l}$. The second part in~\cref{eq:loss} is the estimated controllable bit consumption, including the estimated bits $L_{\text{entropy}}$ for anchor attributes and $L_{\text{hash}}$ for the hash grid. The last term $L_m$ in~\cref{eq:loss} is the masking loss adopted from \cite{Joo} to regularize the adaptive offset masking module. 
$\lambda_e$ and $\lambda_m$ are trade-off hyperparameters used to balance the loss components. Note that we incorporate different techniques or loss items at different iterations to stabilize the training process. Please refer to the supplementary Sec.\textcolor{red}{A} for more details.

For the encoding/decoding process, the binary hash grid $\mathcal{H}$ is first encoded/ decoded using AE with $h_f$. Then, hash feature $\bm{f}^h$ is obtained through interpolation based on $\mathcal{H}$ and $\bm{x}^a$. Once $\bm{f}^h$ is acquired, the context models ${\rm MLP_q}$ and ${\rm MLP_c}$ are then employed to estimate quantization refinement term $\bm{r}$ and parameters of the Gaussian Distribution (\ie, $\bm{\mu}$ and $\bm{\sigma}$) to derive the probability $p(\bm{\hat{f}})$ for entropy encoding/decoding with AE.

\section{Experiments}
In this section, we first present our HAC framework's implementation details and then conduct evaluation experiments to compare with existing 3DGS compression approaches. Additionally, we include ablation studies to demonstrate the effectiveness of each technical component of our method. Finally, we visualize the bit allocation map for better understanding.  

\subsection{Implementation Details}
We implement our HAC based on the Scaffold-GS repository~\cite{scaffold} using the PyTorch framework~\cite{pytorch} and train the model on a single NVIDIA RTX 4090 GPU. 
We increase the dimension of the Scaffold-GS anchor feature $\bm{f}^a$ (\ie, $D^a$) to 50, and disable its feature bank as we found it may lead to unstable training. For the hash grid $\mathcal{H}$, we utilize a mixed 3D-2D structured binary hash grid, with 12 levels of 3D embeddings ranging from 16 to 512 resolutions, and 4 levels of 2D embeddings ranging from 128 to 1024 resolutions. The maximum hash table sizes are $2^{13}$ and $2^{15}$ for the 3D and 2D grids, respectively, both with a feature dimension of $D^h=4$. We set $\lambda_m$ to $5e-4$, and change $\lambda_e$ from $5e-4$ to $4e-3$ for variable bitrates. We set $Q_0$ as $1$, $0.001$ and $0.2$ for $\bm{f}^a$, $\bm{l}$ and $\bm{o}$, respectively. We combine ${\rm MLP_q}$ and ${\rm MLP_c}$ to a single 3-layer MLP with ReLU activation.


\begin{table}[t] \scriptsize
    \centering
    \setlength\tabcolsep{1.4pt}
    \caption{Quantitative results. 3DGS~\cite{3DGS} and Scaffold-GS~\cite{scaffold} are two baselines. Approaches in the middle chunk are designed for 3DGS compression.
    For our approach, we give two results of different size and fidelity tradeoffs by adjusting $\lambda_e$. 
    A smaller $\lambda_e$ results in a larger size but improved fidelity, and vice versa. 
    The best and 2nd best results are highlighted in \colorbox{red!25}{red} and \colorbox{yellow!25}{yellow} cells. The size is measured in MB.
    }
    \begin{tabular}{ll|cccc|cccc|cccc}
        \toprule
        \multicolumn{2}{l|}{\textbf{Datasets}}          & \multicolumn{4}{c|}{\textbf{Synthetic-NeRF~\cite{NeRF}}} & \multicolumn{4}{c|}{\textbf{Mip-NeRF360~\cite{mip360}}} & \multicolumn{4}{c}{\textbf{Tank\&Temples~\cite{tant}}} \\
        \multicolumn{2}{l|}{\textbf{methods}} & psnr$\uparrow$    & ssim$\uparrow$   & lpips$\downarrow$ & size$\downarrow$   & psnr$\uparrow$    & ssim$\uparrow$   & lpips$\downarrow$ & size$\downarrow$   & psnr$\uparrow$   & ssim$\uparrow$   & lpips$\downarrow$ & size$\downarrow$   \\
        \bottomrule
        \multicolumn{2}{l|}{\textbf{3DGS~\cite{3DGS}}}&\cellcolor{red!25}{33.80}&\cellcolor{red!25}{0.970}&\cellcolor{red!25}{0.031}&68.46&27.49&\cellcolor{red!25}{0.813}&\cellcolor{red!25}{0.222}&744.7&23.69&0.844&\cellcolor{yellow!25}{0.178}&431.0    \\
        \multicolumn{2}{l|}{\textbf{Scaffold-GS~\cite{scaffold}}}&33.41&0.966&0.035&19.36&27.50&0.806&0.252&253.9&23.96&\cellcolor{red!25}{0.853}&\cellcolor{red!25}{0.177}&86.50    \\  \hline
        \multicolumn{2}{l|}{\textbf{Lee~\etal~\cite{Joo}}}&33.33&\cellcolor{yellow!25}{0.968}&0.034&5.54&27.08&0.798&0.247&48.80&23.32&0.831&0.201&39.43    \\  
        \multicolumn{2}{l|}{\textbf{Compressed3D~\cite{Simon}}}&32.94&0.967&\cellcolor{yellow!25}{0.033}&3.68&26.98&0.801&0.238&28.80&23.32&0.832&0.194&17.28    \\  
        \multicolumn{2}{l|}{\textbf{EAGLES\cite{Sharath}}}&32.54&0.965&0.039&5.74&27.15&0.808&0.238&68.89&23.41&0.840&0.200&34.00    \\
        \multicolumn{2}{l|}{\textbf{LightGaussian~\cite{zhiwen}}}&32.73&0.965&0.037&7.84&27.00&0.799&0.249&44.54&22.83&0.822&0.242&22.43    \\  
        \multicolumn{2}{l|}{\textbf{Morgen.~\etal~\cite{Wieland}}}&31.05&0.955&0.047&2.20&26.01&0.772&0.259&23.90&22.78&0.817&0.211&13.05    \\
        \multicolumn{2}{l|}{\textbf{Navaneet~\etal~\cite{KLNavaneet}}}&33.09&0.967&0.036&4.42&27.16&0.808&\cellcolor{yellow!25}{0.228}&50.30&23.47&0.840&0.188&27.97    \\  \hline
        \multicolumn{2}{l|}{\textbf{Ours-lowrate}}&33.24&0.967&0.037&\cellcolor{red!25}{1.18}&\cellcolor{yellow!25}{27.53}&0.807&0.238&\cellcolor{red!25}{15.26}&\cellcolor{yellow!25}{24.04}&\cellcolor{yellow!25}{0.846}&0.187&\cellcolor{red!25}{8.10}    \\  
        \multicolumn{2}{l|}{\textbf{Ours-highrate}}&\cellcolor{yellow!25}{33.71}&\cellcolor{yellow!25}{0.968}&0.034&\cellcolor{yellow!25}{1.86}&\cellcolor{red!25}{27.77}&\cellcolor{yellow!25}{0.811}&0.230&\cellcolor{yellow!25}{21.87}&\cellcolor{red!25}{24.40}&\cellcolor{red!25}{0.853}&\cellcolor{red!25}{0.177}&\cellcolor{yellow!25}{11.24}    \\  \hline
        \toprule
    \end{tabular}
    \centering
    \begin{tabular}{ll|cccc|cccc}
        \toprule
        \multicolumn{2}{l|}{\textbf{Datasets}} & \multicolumn{4}{c|}{\textbf{DeepBlending~\cite{deepblending}}} & \multicolumn{4}{c}{\textbf{BungeeNeRF~\cite{BungeeNeRF}}} \\
        \multicolumn{2}{l|}{\textbf{methods}} & psnr$\uparrow$    & ssim$\uparrow$   & lpips$\downarrow$ & size$\downarrow$   & psnr$\uparrow$    & ssim$\uparrow$   & lpips$\downarrow$ & size$\downarrow$   \\
        \bottomrule
        \multicolumn{2}{l|}{\textbf{3DGS~\cite{3DGS}}}&29.42&0.899&\cellcolor{red!25}{0.247}&663.9&24.87&0.841&\cellcolor{red!25}{0.205}&1616    \\
        \multicolumn{2}{l|}{\textbf{Scaffold-GS~\cite{scaffold}}}&\cellcolor{yellow!25}{30.21}&\cellcolor{yellow!25}{0.906}&0.254&66.00&\cellcolor{yellow!25}{26.62}&\cellcolor{yellow!25}{0.865}&0.241&183.0    \\  \hline
        \multicolumn{2}{l|}{\textbf{Lee~\etal~\cite{Joo}}}&29.79&0.901&0.258&43.21&23.36&0.788&0.251&82.60    \\  
        \multicolumn{2}{l|}{\textbf{Compressed3D~\cite{Simon}}}&29.38&0.898&0.253&25.30&24.13&0.802&0.245&55.79    \\  
        \multicolumn{2}{l|}{\textbf{EAGLES\cite{Sharath}}}&29.91&\cellcolor{red!25}{0.910}&\cellcolor{yellow!25}{0.250}&62.00&25.24&0.843&0.221&117.1    \\
        \multicolumn{2}{l|}{\textbf{LightGaussian~\cite{zhiwen}}}&27.01&0.872&0.308&33.94&24.52&0.825&0.255&87.28    \\  
        \multicolumn{2}{l|}{\textbf{Morgen.~\etal~\cite{Wieland}}}&28.92&0.891&0.276&8.40&$-$&$-$&$-$&$-$    \\
        \multicolumn{2}{l|}{\textbf{Navaneet~\etal~\cite{KLNavaneet}}}&29.75&0.903&\cellcolor{red!25}{0.247}&42.77&24.63&0.823&0.239&104.3    \\  \hline
        \multicolumn{2}{l|}{\textbf{Ours-lowrate}}&29.98&0.902&0.269&\cellcolor{red!25}{4.35}&26.48&0.845&0.250&\cellcolor{red!25}{18.49}    \\  
        \multicolumn{2}{l|}{\textbf{Ours-highrate}}&\cellcolor{red!25}{30.34}&\cellcolor{yellow!25}{0.906}&0.258&\cellcolor{yellow!25}{6.35}&\cellcolor{red!25}{27.08}&\cellcolor{red!25}{0.872}&\cellcolor{yellow!25}{0.209}&\cellcolor{yellow!25}{29.72}    \\  \hline
        \toprule
    \end{tabular}
    \label{tab:main_quantitative}
\end{table}

\subsection{Experiment Evaluation}
\textbf{Baselines}. We compare our HAC with existing 3DGS compression approaches. Notably,~\cite{Joo, Simon, KLNavaneet, zhiwen} mainly adopt codebook-based or parameter pruning strategies, while Scaffold-GS~\cite{scaffold} explores Gaussian relations for compact representation. Additionally, EAGLES~\cite{Sharath} and Morgenstern~\etal~\cite{Wieland} employ non-contextual entropy constraints and dimension collapse techniques, respectively.

\noindent\textbf{Datasets}. We follow Scaffold-GS to perform evaluations on multiple datasets, including the small-scale Synthetic-NeRF~\cite{NeRF} and the four large-scale real-scene datasets: BungeeNeRF~\cite{BungeeNeRF}, DeepBlending~\cite{deepblending}, Mip-NeRF360~\cite{mip360}, and Tanks\&Temples~\cite{tant}. Note that we evaluate the entire 9 scenes from Mip-NeRF360 dataset~\cite{mip360}. Covering diverse scenarios, these datasets allow us to comprehensively demonstrate the effectiveness of our approach.

\noindent\textbf{Metrics}. To comprehensively evaluate compression Rate-Distortion (RD) performance, we calculate relative rate (size) change of our approach over others under a similar fidelity. 
Note that BD-rate~\cite{BD_rate} is incalculable as other methods can typically only output a single rate, while four are needed for its calculation.

\begin{figure}[t]
    \centering
    \includegraphics[width=1.00\linewidth]{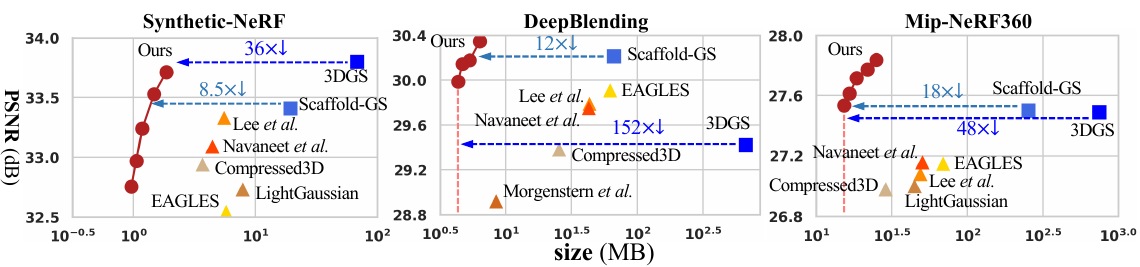}
    \caption{RD curves for quantitative comparisons. We vary $\lambda_e$ to achieve variable bitrates. Note that ${\rm log_{10}}$ scale is used for x-axis for better visualization.
    }
    \label{fig:main_quantitative_curve}
\end{figure}

\begin{figure}[t]
    \centering
    \includegraphics[width=1.00\linewidth]{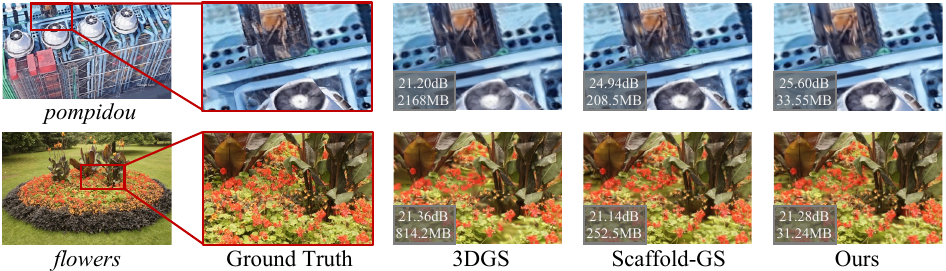}
    \caption{Qualitative comparisons of ``pompidou'' from BungeeNeRF~\cite{BungeeNeRF} and ``flowers'' from Mip-NeRF360~\cite{mip360}. PSNR and size results are given at lower-left.}
    \label{fig:main_qualitative}
\end{figure}

\noindent\textbf{Results}. Quantitative results are shown in~\cref{tab:main_quantitative} and~\cref{fig:main_quantitative_curve}, the qualitative outputs are presented in~\cref{fig:main_qualitative}. Please refer to the supplementary Sec.\textcolor{red}{C} for detailed metrics of each scene. Our HAC has demonstrated significant size reduction of over $75\times$ compared to the vanilla 3DGS~\cite{3DGS} with even improved fidelity. The size reduction also exceeds $11\times$ over the base model Scaffold-GS~\cite{scaffold}. Notably, our highest fidelity surpasses Scaffold-GS, primarily due to two factors: 1) the entropy loss effectively regularizes the model to prevent overfitting, and 2) we increase the dimension of the anchor feature (\ie, $D^a$) to 50, resulting in a larger model volume. Although other compression approaches (mid chunk) can reduce the model size by primarily using pruning and codebooks, they still exhibit significant spatial redundancy. Specifically, Morgenstern~\cite{Wieland} achieves a comparably small size, but significantly sacrifices fidelity due to the dimension collapsing.

\noindent\textbf{Bitstream}. Our bitstream consists of five components: anchor attributes $\mathcal{A}$ (comprising $\bm{f}^a$, $\bm{l}$ and $\bm{o}$), binary hash grid $\mathcal{H}$, offset masks, anchor locations $\bm{x}^a$ and MLPs. Among them, $\mathcal{A}$ is encoded using entropy codec AE~\cite{AE} with estimated probabilities from HAC. It accounts for the dominant portion of the storage. The hash grid $\mathcal{H}$ and the masks are binary data and are encoded by AE using the respective occurrence frequency. The last two components are stored directly in 16 and 32 bits, respectively. When analyzing the bit allocation of each component, they are 14.90MB (8.76MB, 2.52MB, 3.62MB), 0.15MB, 0.52MB, 2.77MB, and 0.16MB for these five components on the most challenging BungeeNeRF dataset~\cite{BungeeNeRF} with $\lambda_e=4e-3$. With scenes become simpler, the storage share of $\mathcal{A}$ decreases as the value distribution become easier to predict.

\begin{figure}[t]
    \centering
    \includegraphics[width=1.00\linewidth]{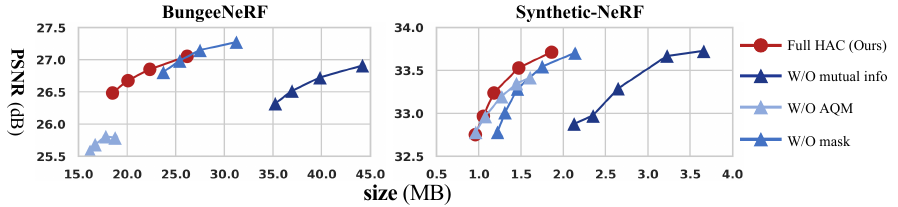}
    \caption{Ablations of different components in HAC. We vary $\lambda_e$ for variable rates.}
    \label{fig:ablation}
\end{figure}

\subsection{Ablation Study}
In this subsection, we conduct ablation studies to demonstrate effectiveness of each technical component. We conduct experiments on both the most challenging large-scale BungeeNeRF dataset~\cite{BungeeNeRF} and the small-scale Synthetic-NeRF dataset~\cite{NeRF} to support convincing and solid results.
We assess the effectiveness of individual technical components by disabling either of the following: 1) mutual information from the hash grid, 2) the adaptive quantization module, 3) adaptive offset masking. The results are presented in~\cref{fig:ablation}.
Firstly, we set the hash grid to all zeros to eliminate mutual information. This leads to a degradation of conditional probability from $p(\mathcal{A}|\bm{f}^h)$ to $p(\mathcal{A})$, which indicates that probability of $\bm{\hat{f}}$ can only be estimated by the statistic $\mu$ and $\sigma$ from the left part of~\cref{fig:hash_gaussian_mask}. Consequently, the bit consumption drastically increases as the probability can no longer be accurately estimated.
Regarding the latter two components, they contribute from different perspectives. Disabling AQM (we remove ${\bm{r}}$ while retaining $Q_0$ to ensure a necessary decimal quantization step) results in a significant drop in fidelity, especially in more complex scenes or at higher rates, as $\bm{\hat{f}}$ fails to retain sufficient information for rendering after quantization. Differently, offset masking can achieve remarkable rate savings in simpler scenes or lower rate segments due to more significant positional redundancy in Gaussians.
Overall, all three components provide a worthwhile tradeoff for improved RD performance.

\begin{figure}[t]
    \centering
    \includegraphics[width=1.00\linewidth]{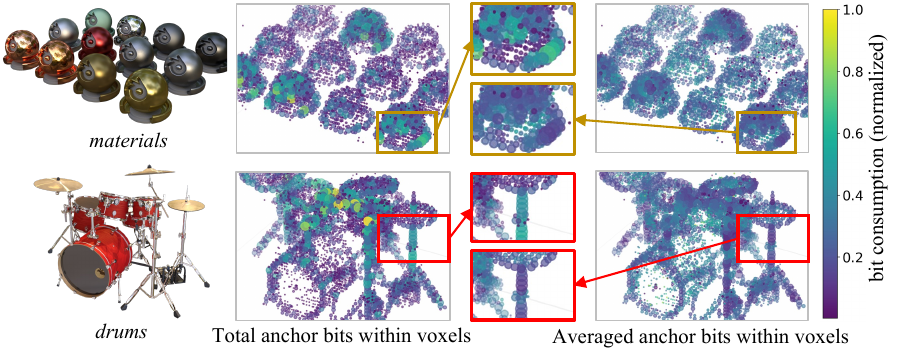}
    \caption{Visualization of bit allocation maps for the scenes ``materials'' and ``drums'' on Synthetic-NeRF dataset~\cite{NeRF}. The 3D space is voxelized, with each voxel represented by a ball and the radius of a ball indicating the number of anchors in the voxel. 
    For the 2nd column, 
    the color of a ball indicates the \textit{total} bit consumption of all anchors in the voxel, while for the 4th column, 
    the color represents the \textit{averaged} bit consumption per anchor within a voxel. The 3rd column gives zoom-in views. It shows more anchors are allocated to important regions while the bit consumption for each anchor is smooth.}
    \label{fig:bit_allocation}
\end{figure}

\subsection{Visualization of Bit Allocation}

While HAC measures the parameters' bit consumption, we are interested in the bit allocation across different local areas in the space. 
In~\cref{fig:bit_allocation}, we utilize scenes in Synthetic-NeRF dataset~\cite{NeRF} for visualization, and represent bit allocation conditions by voxelized colored balls. As observed from the 2nd column of visualized sub-figures, the model tends to allocate more total bits to areas with complex appearances or sharp edges. For instance, specular objects in ``materials'' and instrument stands in ``drums'' exhibit higher total bit consumption due to the complex textures. The analysis of the 4th column from an averaging viewpoint reveals varied trends in bit consumption per anchor. In high bit-consumption voxels, creating more anchors for precise modeling averages the bit per anchor, smoothing or reducing bit consumption for each. This aligns with our assumption that anchors demonstrate inherent consistency in the 3D space where nearby anchors exhibit similar values of attributes, making it easier for the hash grid to accurately estimate their value probabilities.



\subsection{Training and Execution Time}

\noindent\textbf{Training time}. Use of additional models in HAC results in increased training time, approximately $0.9\times$ longer than Scaffold-GS. For the challenging BungeeNeRF dataset~\cite{BungeeNeRF}, the training times are 38.2 $m$ for 3DGS~\cite{3DGS}, 15.1 $m$ for Scaffold-GS~\cite{scaffold} and 27.6 $m$ for HAC. For the small-scale Synthetic-NeRF dataset~\cite{NeRF}, training times are 3.4 $m$, 4.4 $m$ and 9.0 $m$, respectively. This increase of training time in our model over Scaffold-GS is our main limitation, but it is still fast.

\noindent\textbf{Coding time}. The encoding/decoding process takes approximately 0.87 seconds and 26.7 seconds on Synthetic-NeRF and BungeeNeRF dataset under $\lambda_e=4e-3$, respectively. The dominant time consumption occurs during Codec execution of AE on the CPU (over $90\%$), as we only use a single thread.

\noindent\textbf{Inference time}. The inference process benefits from the design of context modeling, allowing for the removal of the hash grid once $\mathcal{A}$ is decoded. Consequently, no additional operations are required during rendering, resulting in a similar FPS with Scaffold-GS. 
The rendering FPS are 75, 232 and 283 for 3DGS, Scaffold-GS and HAC on BungeeNeRF, and 401, 326 and 341 on Synthetic-NeRF. The improved FPS of our model compared to Scaffold-GS is likely due to the pruning of invalid Gaussians/anchors, which in turn facilitates faster rendering.

\section{Conclusion}

We pioneered an investigation into the relationship between unorganized and sparse Gaussians (or anchors) and well-structured hash grids, leveraging their mutual information for compact 3DGS representations. Our Hash-grid Assisted Context (HAC) framework has achieved SoTA compression performance with remarkable leading over concurrent works. Extensive experiments have demonstrated the effectiveness of our HAC and its technical components. Overall, our work has successfully mitigated the major challenging of 3DGS models, \ie, large storage, enabling its adoption in large-scale scenes and diverse devices.

\section*{Acknowledgement}
The paper is supported in part by The National Natural Science Foundation of China (No. 62325109, U21B2013). 

\noindent MH is supported by funding from The Australian Research Council Discovery Program DP230101176.


\bibliographystyle{splncs04}
\bibliography{main}
\clearpage
\setcounter{page}{1}
\centerline{\noindent\textbf{{\Large --Supplementary Material--}}}

\renewcommand\thesection{\Alph{section}}
\renewcommand\thetable{\Alph{table}}
\renewcommand\thefigure{\Alph{figure}}
\setcounter{section}{0}
\setcounter{table}{0}
\setcounter{figure}{0}

\begin{abstract}
  
  This is the supplementary material for our paper. Herein, we offer more details of implementations, an extra experiment, quantitative per-scene results across all datasets, and a comprehensive notation table.

\end{abstract}

\begin{figure}[h]
    \centering
    \includegraphics[width=0.95\linewidth]{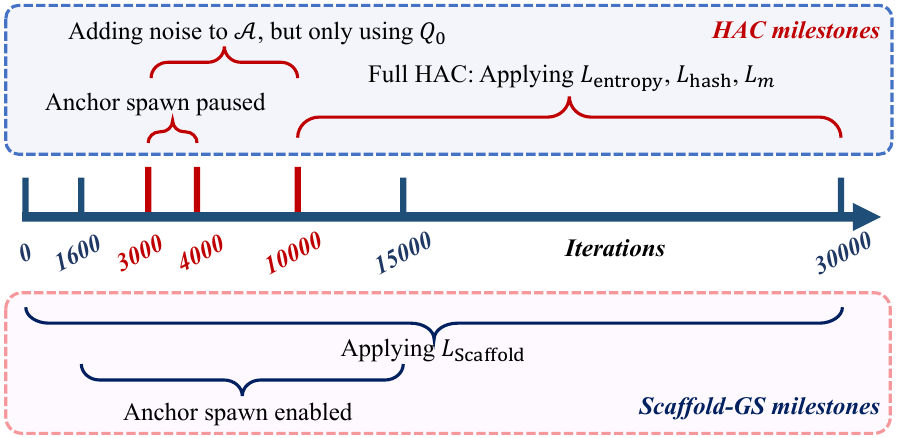}
    \caption{Detailed training process of our HAC model. We use the \textcolor{supple_blue}{blue box} to indicate the training process related to our model, while using the \textcolor{supple_pink}{pink box} for Scaffold-GS~\cite{scaffold}.}
    \label{fig:supple_training}
\end{figure}

\section{More Implementation Details}

\subsection{Training Process}
We provide a detailed overview of the training process for our HAC framework, as illustrated in~\cref{fig:supple_training}. 

\noindent\textbf{During the initial 3000 iterations}, no additional techniques are applied to impact the original training process of Scaffold-GS~\cite{scaffold}, ensuring a stable start of the anchor attribute training and anchor spawning.

\noindent\textbf{From iteration 3000 to 10000}, we introduce ``adding noise'' operations to anchor attributes $\mathcal{A}$, which allows the model to adapt to the quantization process. Note that, in this stage, we only apply $Q_0$ for quantization without using $\bm{r}$ for refinement, therefore, we do not need the hash grid.
Specifically, we pause the anchor spawning process between iterations 3000 and 4000 for a transitional period, as the sudden introduction of quantization may introduce instability to the spawning process. Once the parameters are fitted to the quantization after iteration 4000, we re-enable the spawning process. Note that we do not incorporate the hash grid in this stage (\ie, before iteration 10000) because we want to provide a transition for the anchor attributes and the spawning process to fit the quantization operation, enabling a more stable training process when the hash grid is incorporated in the further iterations.

\noindent\textbf{After iteration 10000}, assuming the 3D model is adequately fitted to the quantization, we fully integrate our HAC framework to jointly train the binary hash grid. Notably, the bound of the hash grid is determined using the maximum and minimum anchor locations at the 10000th iteration, which are then utilized to normalize anchor locations for interpolation in the hash grid. This comprehensive pipeline ensures a stable training process to reduce the model size via entropy constraints while maintaining a high-quality fidelity.

\subsection{Sampling Strategy}
During training, employing all anchors for entropy training in each iteration could result in prolonged training time and potential out-of-memory (OOM) issues. Therefore, we adopt a sampling strategy that, in each iteration, we only randomly sample and entropy train 5\% of anchors from those used for rendering. This approach ensures faster training speeds while still preserving satisfactory RD performance.

\section{Additional Experiments}

\begin{table}[t]
    \centering
    \caption{Bit allocation among anchor's attributes. We set $\lambda_e=4e-3$. When calculating per-param size, we only consider \textit{valid} anchors that are not pruned.}
    \setlength\tabcolsep{9pt}
    \begin{tabular}{l|ccc|ccc}
    \toprule[1pt]
    \multirow{2}{*}{\textbf{Dataset}} & \multicolumn{3}{c|}{Total size (MB)} & \multicolumn{3}{c}{Per-param size (bit)} \\ \cline{2-7}
     & $\bm{f}^a$ & $\bm{l}$ & $\bm{o}$ & $\bm{f}^a$ & $\bm{l}$ & $\bm{o}$  \\ \hline
    \textbf{Bungee-NeRF}~\cite{BungeeNeRF} & 8.76 & 2.53 & 3.62 & 3.03 & 7.27 & 4.56 \\
    \textbf{Synthetic-NeRF}~\cite{NeRF} & 0.31 & 0.09 & 0.12 & 1.33 & 3.58 & 3.76 \\
    \toprule[1pt]
    \end{tabular}
    \label{tab:bit_allocation_flo}
\end{table}

We investigate bit allocation among the anchor's three attributes, as depicted in~\cref{tab:bit_allocation_flo}. When viewing from the total size, feature $\bm{f}^a$ contributes the most due to its highest dimensionality. However, as it needs to be inputted into MLPs to extract Gaussian attributes, it exhibits the most significant dimensional redundancy, resulting in the smallest per-parameter bit. Conversely, this is not the case for scaling $\bm{l}$ and offsets $\bm{o}$, which are directly used for rendering, making much fewer dimensional redundancies. Additionally, as $\bm{l}$ and $\bm{o}$ are always of higher decimal precision, their value distributions are more difficult to accurately predict, resulting higher per-parameter bit consumption.

\section{Quantitative Results of Each Scene}

\subsection{Detailed Results of Our HAC Framework}

\noindent Detailed per-scene results of Synthetic-NeRF dataset~\cite{NeRF} are given in~\cref{tab:supple_blender}.

\noindent Detailed per-scene results of Mip-NeRF360 dataset~\cite{mip360} are given in~\cref{tab:supple_mip360}.

\noindent Detailed per-scene results of Tank\&Temples dataset~\cite{tant} are given in~\cref{tab:supple_tandt}.

\noindent Detailed per-scene results of DeepBlending dataset~\cite{deepblending} are given in~\cref{tab:supple_blending}.

\noindent Detailed per-scene results of BungeeNeRF dataset~\cite{BungeeNeRF} are given in~\cref{tab:supple_bungee}.

\subsection{Detailed Results of the Base Models}

\noindent We also give detailed per-scene results of all datasets of our two base models 3DGS~\cite{3DGS} and Scaffold-GS~\cite{scaffold} in~\cref{tab:supple_3DGS} and~\cref{tab:supple_scaffolld}, respectively.

\section{Notation Table}
Please refer to~\cref{tab:notation_table} for detailed notation explanations.

\clearpage
\begin{table}[t]
\centering
\setlength\tabcolsep{6pt}  
\renewcommand{\arraystretch}{0.95}  
\caption{Per-scene results of Synthetic-NeRF dataset~\cite{NeRF} of our approach.}
\begin{tabular}{c|c|cccc}
\toprule[2pt]
$\lambda_e$ & Scenes & PSNR$\uparrow$ & SSIM$\uparrow$ & LPIPS$\downarrow$ & SIZE$\downarrow$ \\
\toprule[1pt]
\multirow{9}{*}{0.004} & chair & 34.02 & 0.981 & 0.018 & 0.82 \\
 & drums & 26.20 & 0.950 & 0.044 & 1.23 \\
 & ficus & 34.27 & 0.983 & 0.016 & 0.71 \\
 & hotdog & 36.44 & 0.979 & 0.033 & 0.51 \\
 & lego & 34.25 & 0.976 & 0.027 & 0.97 \\
 & materials & 30.20 & 0.959 & 0.045 & 1.07 \\
 & mic & 35.39 & 0.989 & 0.011 & 0.55 \\
 & ship & 31.24 & 0.902 & 0.124 & 1.82 \\
 & \textbf{AVG} & \textbf{32.75} & \textbf{0.965} & \textbf{0.040} & \textbf{0.96} \\
 \midrule
\multirow{9}{*}{0.003} & chair & 34.33 & 0.982 & 0.017 & 0.89 \\
 & drums & 26.26 & 0.951 & 0.043 & 1.42 \\
 & ficus & 34.57 & 0.984 & 0.015 & 0.82 \\
 & hotdog & 36.70 & 0.980 & 0.031 & 0.54 \\
 & lego & 34.65 & 0.977 & 0.024 & 1.07 \\
 & materials & 30.29 & 0.960 & 0.043 & 1.20 \\
 & mic & 35.62 & 0.990 & 0.010 & 0.62 \\
 & ship & 31.32 & 0.903 & 0.121 & 1.86 \\
 & \textbf{AVG} & \textbf{32.97} & \textbf{0.966} & \textbf{0.038} & \textbf{1.05} \\
 \midrule
\multirow{9}{*}{0.002} & chair & 34.73 & 0.984 & 0.016 & 1.03 \\
 & drums & 26.32 & 0.952 & 0.043 & 1.45 \\
 & ficus & 34.90 & 0.985 & 0.014 & 0.94 \\
 & hotdog & 37.11 & 0.981 & 0.029 & 0.64 \\
 & lego & 35.04 & 0.979 & 0.022 & 1.25 \\
 & materials & 30.53 & 0.961 & 0.041 & 1.45 \\
 & mic & 35.92 & 0.990 & 0.010 & 0.67 \\
 & ship & 31.38 & 0.903 & 0.119 & 1.99 \\
 & \textbf{AVG} & \textbf{33.24} & \textbf{0.967} & \textbf{0.037} & \textbf{1.18} \\
 \midrule
\multirow{9}{*}{0.001} & chair & 35.21 & 0.985 & 0.014 & 1.32 \\
 & drums & 26.38 & 0.952 & 0.041 & 1.95 \\
 & ficus & 35.37 & 0.986 & 0.013 & 1.20 \\
 & hotdog & 37.47 & 0.983 & 0.026 & 0.79 \\
 & lego & 35.51 & 0.981 & 0.019 & 1.61 \\
 & materials & 30.58 & 0.961 & 0.040 & 1.62 \\
 & mic & 36.25 & 0.991 & 0.009 & 0.81 \\
 & ship & 31.48 & 0.904 & 0.116 & 2.50 \\
 & \textbf{AVG} & \textbf{33.53} & \textbf{0.968} & \textbf{0.035} & \textbf{1.47} \\
 \midrule
\multirow{9}{*}{0.0005} & chair & 35.49 & 0.986 & 0.013 & 1.67 \\
 & drums & 26.45 & 0.952 & 0.041 & 2.32 \\
 & ficus & 35.30 & 0.986 & 0.013 & 1.53 \\
 & hotdog & 37.87 & 0.984 & 0.024 & 0.97 \\
 & lego & 35.67 & 0.981 & 0.019 & 1.90 \\
 & materials & 30.70 & 0.962 & 0.039 & 2.07 \\
 & mic & 36.71 & 0.992 & 0.008 & 1.01 \\
 & ship & 31.52 & 0.904 & 0.115 & 3.39 \\
 & \textbf{AVG} & \textbf{33.71} & \textbf{0.968} & \textbf{0.034} & \textbf{1.86} \\
 \toprule[2pt]
\end{tabular}
\label{tab:supple_blender}
\end{table}

\clearpage
\begin{table}[t]
\centering
\setlength\tabcolsep{6pt}  
\renewcommand{\arraystretch}{0.87}  
\caption{Per-scene results of Mip-NeRF360 dataset~\cite{mip360} of our approach.}
\begin{tabular}{c|c|cccc}
\toprule[2pt]
$\lambda_e$ & Scenes & PSNR$\uparrow$ & SSIM$\uparrow$ & LPIPS$\downarrow$ & SIZE$\downarrow$ \\
\toprule[1pt]
\multirow{10}{*}{0.004} & bicycle & 25.05 & 0.742 & 0.264 & 27.54 \\
 & garden & 27.28 & 0.842 & 0.151 & 22.69 \\
 & stump & 26.58 & 0.762 & 0.269 & 18.11 \\
 & room & 31.55 & 0.921 & 0.208 & 5.53 \\
 & counter & 29.35 & 0.911 & 0.195 & 7.26 \\
 & kitchen & 31.16 & 0.923 & 0.131 & 8.05 \\
 & bonsai & 32.28 & 0.942 & 0.189 & 8.56 \\
 & flower & 21.26 & 0.572 & 0.381 & 19.59 \\
 & treehill & 23.30 & 0.645 & 0.356 & 20.04 \\
 & \textbf{AVG} & \textbf{27.53} & \textbf{0.807} & \textbf{0.238} & \textbf{15.26} \\
 \midrule
\multirow{10}{*}{0.003} & bicycle & 25.05 & 0.742 & 0.261 & 30.02 \\
 & garden & 27.36 & 0.844 & 0.148 & 24.62 \\
 & stump & 26.64 & 0.763 & 0.265 & 19.85 \\
 & room & 31.71 & 0.922 & 0.206 & 5.72 \\
 & counter & 29.54 & 0.913 & 0.191 & 7.93 \\
 & kitchen & 31.22 & 0.925 & 0.128 & 8.84 \\
 & bonsai & 32.50 & 0.944 & 0.186 & 9.40 \\
 & flower & 21.26 & 0.571 & 0.383 & 20.67 \\
 & treehill & 23.26 & 0.645 & 0.356 & 22.08 \\
 & \textbf{AVG} & \textbf{27.62} & \textbf{0.808} & \textbf{0.236} & \textbf{16.57} \\
 \midrule
\multirow{10}{*}{0.002} & bicycle & 25.10 & 0.742 & 0.262 & 33.14 \\
 & garden & 27.43 & 0.847 & 0.143 & 27.52 \\
 & stump & 26.59 & 0.761 & 0.268 & 21.75 \\
 & room & 31.87 & 0.925 & 0.201 & 6.47 \\
 & counter & 29.65 & 0.915 & 0.189 & 8.88 \\
 & kitchen & 31.46 & 0.928 & 0.125 & 10.05 \\
 & bonsai & 32.70 & 0.945 & 0.184 & 10.51 \\
 & flower & 21.32 & 0.576 & 0.377 & 23.73 \\
 & treehill & 23.34 & 0.647 & 0.350 & 24.83 \\
 & \textbf{AVG} & \textbf{27.72} & \textbf{0.809} & \textbf{0.233} & \textbf{18.54} \\
 \midrule
\multirow{10}{*}{0.001} & bicycle & 25.11 & 0.742 & 0.259 & 39.15 \\
 & garden & 27.46 & 0.849 & 0.139 & 32.17 \\
 & stump & 26.59 & 0.763 & 0.264 & 25.26 \\
 & room & 31.90 & 0.926 & 0.198 & 7.85 \\
 & counter & 29.74 & 0.918 & 0.184 & 10.44 \\
 & kitchen & 31.63 & 0.930 & 0.122 & 12.07 \\
 & bonsai & 32.97 & 0.948 & 0.180 & 12.72 \\
 & flower & 21.27 & 0.575 & 0.377 & 27.55 \\
 & treehill & 23.26 & 0.648 & 0.345 & 29.65 \\
 & \textbf{AVG} & \textbf{27.77} & \textbf{0.811} & \textbf{0.230} & \textbf{21.87} \\
 \midrule
\multirow{10}{*}{0.0005} & bicycle & 25.05 & 0.742 & 0.258 & 44.01 \\
 & garden & 27.50 & 0.850 & 0.139 & 36.27 \\
 & stump & 26.57 & 0.762 & 0.264 & 28.93 \\
 & room & 32.19 & 0.929 & 0.194 & 9.16 \\
 & counter & 29.75 & 0.918 & 0.185 & 12.22 \\
 & kitchen & 31.81 & 0.931 & 0.120 & 13.96 \\
 & bonsai & 33.16 & 0.949 & 0.178 & 14.90 \\
 & flower & 21.28 & 0.575 & 0.376 & 31.24 \\
 & treehill & 23.22 & 0.646 & 0.346 & 34.42 \\
 & \textbf{AVG} & \textbf{27.83} & \textbf{0.811} & \textbf{0.229} & \textbf{25.01} \\
 \toprule[2pt]
\end{tabular}
\label{tab:supple_mip360}
\end{table}

\clearpage
\begin{table}[t]
\centering
\setlength\tabcolsep{6pt}  
\renewcommand{\arraystretch}{1.00}  
\caption{Per-scene results of Tank\&Temples dataset~\cite{tant} of our approach.}
\begin{tabular}{c|c|cccc}
\toprule[2pt]
$\lambda_e$ & Scenes & PSNR$\uparrow$ & SSIM$\uparrow$ & LPIPS$\downarrow$ & SIZE$\downarrow$ \\
\toprule[1pt]
\multirow{3}{*}{0.004} & truck & 25.88 & 0.878 & 0.158 & 9.26 \\
 & train & 22.19 & 0.815 & 0.216 & 6.94 \\
 & \textbf{AVG} & \textbf{24.04} & \textbf{0.846} & \textbf{0.187} & \textbf{8.10} \\
 \midrule
\multirow{3}{*}{0.003} & truck & 25.99 & 0.880 & 0.153 & 9.80 \\
 & train & 22.49 & 0.817 & 0.213 & 7.59 \\
 & \textbf{AVG} & \textbf{24.24} & \textbf{0.849} & \textbf{0.183} & \textbf{8.70} \\
 \midrule
\multirow{3}{*}{0.002} & truck & 25.99 & 0.881 & 0.153 & 11.15 \\
 & train & 22.66 & 0.819 & 0.210 & 8.64 \\
 & \textbf{AVG} & \textbf{24.33} & \textbf{0.850} & \textbf{0.181} & \textbf{9.90} \\
 \midrule
\multirow{3}{*}{0.001} & truck & 26.02 & 0.883 & 0.147 & 12.42 \\
 & train & 22.78 & 0.823 & 0.207 & 10.07 \\
 & \textbf{AVG} & \textbf{24.40} & \textbf{0.853} & \textbf{0.177} & \textbf{11.24} \\
 \midrule
\multirow{3}{*}{0.0005} & truck & 26.00 & 0.883 & 0.146 & 15.12 \\
 & train & 22.49 & 0.823 & 0.206 & 11.19 \\
 & \textbf{AVG} & \textbf{24.25} & \textbf{0.853} & \textbf{0.176} & \textbf{13.16} \\
 \toprule[2pt]
\end{tabular}
\label{tab:supple_tandt}
\end{table}

\begin{table}[t]
\centering
\setlength\tabcolsep{6pt}  
\renewcommand{\arraystretch}{1.00}  
\caption{Per-scene results of DeepBlending dataset~\cite{deepblending} of our approach.}
\begin{tabular}{c|c|cccc}
\toprule[2pt]
$\lambda_e$ & Scenes & PSNR$\uparrow$ & SSIM$\uparrow$ & LPIPS$\downarrow$ & SIZE$\downarrow$ \\
\toprule[1pt]
\multirow{3}{*}{0.004} & playroom & 30.44 & 0.902 & 0.272 & 3.15 \\
 & drjohnson & 29.53 & 0.903 & 0.265 & 5.55 \\
 & \textbf{AVG} & \textbf{29.98} & \textbf{0.902} & \textbf{0.269} & \textbf{4.35} \\
 \midrule
\multirow{3}{*}{0.003} & playroom & 30.61 & 0.903 & 0.269 & 3.66 \\
 & drjohnson & 29.67 & 0.904 & 0.261 & 5.71 \\
 & \textbf{AVG} & \textbf{30.14} & \textbf{0.903} & \textbf{0.265} & \textbf{4.69} \\
 \midrule
\multirow{3}{*}{0.002} & playroom & 30.66 & 0.905 & 0.265 & 4.12 \\
 & drjohnson & 29.69 & 0.905 & 0.258 & 6.51 \\
 & \textbf{AVG} & \textbf{30.17} & \textbf{0.905} & \textbf{0.262} & \textbf{5.32} \\
 \midrule
\multirow{3}{*}{0.001} & playroom & 30.84 & 0.906 & 0.262 & 5.03 \\
 & drjohnson & 29.85 & 0.906 & 0.255 & 7.67 \\
 & \textbf{AVG} & \textbf{30.34} & \textbf{0.906} & \textbf{0.258} & \textbf{6.35} \\
 \midrule
\multirow{3}{*}{0.0005} & playroom & 30.66 & 0.906 & 0.259 & 6.08 \\
 & drjohnson & 29.76 & 0.906 & 0.255 & 9.09 \\
 & \textbf{AVG} & \textbf{30.21} & \textbf{0.906} & \textbf{0.257} & \textbf{7.58} \\
 \toprule[2pt]
\end{tabular}
\label{tab:supple_blending}
\end{table}

\clearpage
\begin{table}[t]
\centering
\setlength\tabcolsep{6pt}  
\renewcommand{\arraystretch}{1.00}  
\caption{Per-scene results of BungeeNeRF dataset~\cite{BungeeNeRF} of our approach.}
\begin{tabular}{c|c|cccc}
\toprule[2pt]
$\lambda_e$ & Scenes & PSNR$\uparrow$ & SSIM$\uparrow$ & LPIPS$\downarrow$ & SIZE$\downarrow$ \\
\toprule[1pt]
\multirow{7}{*}{0.004} & amsterdam & 26.80 & 0.865 & 0.224 & 22.49 \\
 & bilbao & 27.65 & 0.864 & 0.231 & 17.14 \\
 & hollywood & 24.25 & 0.748 & 0.347 & 16.55 \\
 & pompidou & 25.16 & 0.829 & 0.266 & 20.40 \\
 & quebec & 29.33 & 0.918 & 0.192 & 15.06 \\
 & rome & 25.68 & 0.845 & 0.243 & 19.30 \\
 & \textbf{AVG} & \textbf{26.48} & \textbf{0.845} & \textbf{0.250} & \textbf{18.49} \\
 \midrule
\multirow{7}{*}{0.003} & amsterdam & 26.95 & 0.873 & 0.214 & 24.41 \\
 & bilbao & 27.82 & 0.872 & 0.218 & 18.76 \\
 & hollywood & 24.27 & 0.753 & 0.342 & 17.87 \\
 & pompidou & 25.34 & 0.837 & 0.255 & 22.49 \\
 & quebec & 29.67 & 0.924 & 0.185 & 16.15 \\
 & rome & 25.98 & 0.855 & 0.231 & 20.83 \\
 & \textbf{AVG} & \textbf{26.67} & \textbf{0.852} & \textbf{0.241} & \textbf{20.08} \\
 \midrule
\multirow{7}{*}{0.002} & amsterdam & 27.13 & 0.880 & 0.202 & 27.14 \\
 & bilbao & 28.02 & 0.880 & 0.205 & 20.91 \\
 & hollywood & 24.43 & 0.763 & 0.330 & 20.09 \\
 & pompidou & 25.27 & 0.842 & 0.249 & 24.85 \\
 & quebec & 29.98 & 0.929 & 0.175 & 17.90 \\
 & rome & 26.28 & 0.866 & 0.219 & 23.07 \\
 & \textbf{AVG} & \textbf{26.85} & \textbf{0.860} & \textbf{0.230} & \textbf{22.33} \\
 \midrule
\multirow{7}{*}{0.001} & amsterdam & 27.25 & 0.886 & 0.190 & 31.84 \\
 & bilbao & 27.98 & 0.886 & 0.190 & 24.38 \\
 & hollywood & 24.59 & 0.772 & 0.319 & 23.41 \\
 & pompidou & 25.58 & 0.851 & 0.236 & 29.19 \\
 & quebec & 30.30 & 0.934 & 0.163 & 21.23 \\
 & rome & 26.61 & 0.876 & 0.203 & 26.91 \\
 & \textbf{AVG} & \textbf{27.05} & \textbf{0.868} & \textbf{0.217} & \textbf{26.16} \\
 \midrule
\multirow{7}{*}{0.0005} & amsterdam & 27.24 & 0.891 & 0.180 & 36.31 \\
 & bilbao & 28.09 & 0.891 & 0.181 & 27.72 \\
 & hollywood & 24.60 & 0.778 & 0.313 & 26.00 \\
 & pompidou & 25.60 & 0.853 & 0.231 & 33.55 \\
 & quebec & 30.19 & 0.936 & 0.155 & 24.56 \\
 & rome & 26.73 & 0.881 & 0.194 & 30.21 \\
 & \textbf{AVG} & \textbf{27.08} & \textbf{0.872} & \textbf{0.209} & \textbf{29.72} \\
 \toprule[2pt]
\end{tabular}
\label{tab:supple_bungee}
\end{table}

\clearpage
\begin{table}[t]
\centering
\setlength\tabcolsep{6pt}  
\renewcommand{\arraystretch}{1.00}  
\caption{Per-scene results of all evaluated datasets of 3DGS~\cite{3DGS}.}
\begin{tabular}{c|c|cccc}
\toprule[2pt]
Datasets & Scenes & PSNR$\uparrow$ & SSIM$\uparrow$ & LPIPS$\downarrow$ & SIZE$\downarrow$ \\
\toprule[1pt]
\multirow{9}{*}{Synthetic-NeRF} & chair & 35.65 & 0.988 & 0.010 & 115.77 \\
 & drums & 26.28 & 0.955 & 0.037 & 92.87 \\
 & ficus & 35.48 & 0.987 & 0.012 & 64.53 \\
 & hotdog & 38.05 & 0.985 & 0.020 & 43.37 \\
 & lego & 35.98 & 0.982 & 0.017 & 80.53 \\
 & materials & 30.48 & 0.960 & 0.037 & 38.50 \\
 & mic & 36.76 & 0.993 & 0.006 & 48.31 \\
 & ship & 31.73 & 0.907 & 0.107 & 63.82 \\
 & \textbf{AVG} & \textbf{33.80} & \textbf{0.970} & \textbf{0.031} & \textbf{68.46} \\
 \midrule
\multirow{10}{*}{Mip-NeRF360} & bicycle & 25.11 & 0.746 & 0.245 & 1336.45 \\
 & garden & 27.30 & 0.856 & 0.122 & 1327.99 \\
 & stump & 26.66 & 0.770 & 0.242 & 1070.92 \\
 & room & 31.74 & 0.926 & 0.197 & 353.10 \\
 & counter & 29.07 & 0.914 & 0.184 & 277.39 \\
 & kitchen & 31.47 & 0.931 & 0.117 & 414.33 \\
 & bonsai & 32.12 & 0.946 & 0.181 & 295.33 \\
 & flower & 21.36 & 0.588 & 0.360 & 814.24 \\
 & treehill & 22.62 & 0.636 & 0.347 & 812.63 \\
 & \textbf{AVG} & \textbf{27.49} & \textbf{0.813} & \textbf{0.222} & \textbf{744.71} \\
 \midrule
\multirow{3}{*}{Tank\&Temples} & truck & 25.38 & 0.877 & 0.148 & 606.99 \\
 & train & 22.00 & 0.811 & 0.208 & 254.91 \\
 & \textbf{AVG} & \textbf{23.69} & \textbf{0.844} & \textbf{0.178} & \textbf{430.95} \\
 \midrule
\multirow{3}{*}{DeepBlending} & playroom & 29.83 & 0.900 & 0.247 & 551.93 \\
 & drjohnson & 29.02 & 0.898 & 0.247 & 775.91 \\
 & \textbf{AVG} & \textbf{29.42} & \textbf{0.899} & \textbf{0.247} & \textbf{663.92} \\
 \midrule
\multirow{7}{*}{BungeeNeRF} & amsterdam & 26.03 & 0.874 & 0.170 & 1458.14 \\
 & bilbao & 26.35 & 0.864 & 0.191 & 1350.37 \\
 & hollywood & 23.44 & 0.767 & 0.241 & 1601.76 \\
 & pompidou & 21.20 & 0.772 & 0.266 & 2169.21 \\
 & quebec & 28.83 & 0.923 & 0.156 & 1468.76 \\
 & rome & 23.34 & 0.848 & 0.206 & 1649.12 \\
 & \textbf{AVG} & \textbf{24.87} & \textbf{0.841} & \textbf{0.205} & \textbf{1616.23} \\
 \toprule[2pt]
\end{tabular}
\label{tab:supple_3DGS}
\end{table}

\clearpage
\begin{table}[t]
\centering
\setlength\tabcolsep{6pt}  
\renewcommand{\arraystretch}{1.00}  
\caption{Per-scene results of all evaluated datasets of Scaffold-GS~\cite{scaffold}.}
\begin{tabular}{c|c|cccc}
\toprule[2pt]
Datasets & Scenes & PSNR$\uparrow$ & SSIM$\uparrow$ & LPIPS$\downarrow$ & SIZE$\downarrow$ \\
\toprule[1pt]
\multirow{9}{*}{Synthetic-NeRF} & chair & 34.96 & 0.985 & 0.013 & 15.50 \\
 & drums & 26.36 & 0.949 & 0.045 & 26.93 \\
 & ficus & 34.66 & 0.984 & 0.015 & 16.46 \\
 & hotdog & 37.82 & 0.984 & 0.022 & 11.31 \\
 & lego & 35.48 & 0.981 & 0.018 & 19.84 \\
 & materials & 30.37 & 0.958 & 0.043 & 23.12 \\
 & mic & 36.37 & 0.991 & 0.008 & 14.83 \\
 & ship & 31.27 & 0.896 & 0.119 & 26.90 \\
 & \textbf{AVG} & \textbf{33.41} & \textbf{0.966} & \textbf{0.035} & \textbf{19.36} \\
 \midrule
\multirow{10}{*}{Mip-NeRF360} & bicycle & 24.50 & 0.705 & 0.306 & 248.00 \\
 & garden & 27.17 & 0.842 & 0.146 & 271.00 \\
 & stump & 26.27 & 0.784 & 0.284 & 493.00 \\
 & room & 31.93 & 0.925 & 0.202 & 133.00 \\
 & counter & 29.34 & 0.914 & 0.191 & 194.00 \\
 & kitchen & 31.30 & 0.928 & 0.126 & 173.00 \\
 & bonsai & 32.70 & 0.946 & 0.185 & 258.00 \\
 & flower & 21.14 & 0.566 & 0.417 & 253.00 \\
 & treehill & 23.19 & 0.642 & 0.410 & 262.00 \\
 & \textbf{AVG} & \textbf{27.50} & \textbf{0.806} & \textbf{0.252} & \textbf{253.89} \\
 \midrule
\multirow{3}{*}{Tank\&Temples} & truck & 25.77 & 0.883 & 0.147 & 107.00 \\
 & train & 22.15 & 0.822 & 0.206 & 66.00 \\
 & \textbf{AVG} & \textbf{23.96} & \textbf{0.853} & \textbf{0.177} & \textbf{86.50} \\
 \midrule
\multirow{3}{*}{DeepBlending} & playroom & 30.62 & 0.904 & 0.258 & 63.00 \\
 & drjohnson & 29.80 & 0.907 & 0.250 & 69.00 \\
 & \textbf{AVG} & \textbf{30.21} & \textbf{0.906} & \textbf{0.254} & \textbf{66.00} \\
 \midrule
\multirow{7}{*}{BungeeNeRF} & amsterdam & 27.16 & 0.898 & 0.188 & 223.00 \\
 & bilbao & 26.60 & 0.857 & 0.257 & 178.00 \\
 & hollywood & 24.49 & 0.787 & 0.318 & 155.00 \\
 & pompidou & 24.94 & 0.839 & 0.271 & 209.00 \\
 & quebec & 30.28 & 0.936 & 0.190 & 159.00 \\
 & rome & 26.23 & 0.873 & 0.225 & 174.00 \\
 & \textbf{AVG} & \textbf{26.62} & \textbf{0.865} & \textbf{0.241} & \textbf{183.00} \\
 \toprule[2pt]
\end{tabular}
\label{tab:supple_scaffolld}
\end{table}

\clearpage

\begin{table} 
    \centering
    \renewcommand{\arraystretch}{0.96}  
    \caption{Notation Table. With slight abuse of notation, we use $L_{3d}$ and $L_{2d}$ to represent the number of levels of the 3D and 2D part of the hash grid, respectively.}
    \begin{tabular}{cll}
    \toprule[2pt]
      \textbf{Notation} & \textbf{Shape}   & \textbf{Definition} \\
    \midrule[1pt]
        $\bm{x}$ &$\mathbb{R}^{3}$&A random 3D point\\
        $\bm{\mu}$ &$\mathbb{R}^{3}$&Location of Gaussians in 3DGS~\cite{3DGS}\\
        $\bm{\Sigma}$ &$\mathbb{R}^{3\times3}$&Covariance matrix of Gaussians\\
        $\bm{S}$ &$\mathbb{R}^{3\times3}$&Scale matrix of Gaussians\\
        $\bm{R}$ &$\mathbb{R}^{3\times3}$&Rotation matrix of Gaussians\\
        $\alpha$ &$\mathbb{R}^1$&Opacity of Gaussians after 2D projection\\
        $\bm{c}$ &$\mathbb{R}^{3}$&View-dependent color of Gaussians\\
        $I$ &&Number of Gaussians contributed to the rendering\\
        $\bm{C}$ &$\mathbb{R}^{3}$&The obtained pixel value after rendering\\
        \cline{1-3}
        $\bm{x}^a$ &$\mathbb{R}^{3}$&Anchor location \\
        $\bm{f}^a$ &$\mathbb{R}^{D^a}$&Feature of the anchor \\
        $\bm{l}$ &$\mathbb{R}^{6}$&Scaling of the anchor \\
        $\bm{o}$ &$\mathbb{R}^{3K}$&Offsets of the anchor \\
        $\mathcal{A}$ &&The set of anchor's attributes including \{$\bm{f}^a$, $\bm{l}$, $\bm{o}$\} \\
        $D^a$ &&Dimension of $\bm{f}^a$\\
        $K$ &&Number of offsets per anchor \\
        \cline{1-3}
        $\mathcal{H}$ &&A 3D-2D mixed binary hash grid \\
        $T$ &&Table size of the hash grid at each level\\
        $L$ &&Number of levels of the hash grid\\
        $D^h$ &&Dimension of the vectors of the hash grid\\
        $\bm{\theta}$ &$\mathbb{R}^{D^h}$&A vector of the hash grid \\
        $\bm{f}^h$ &$\mathbb{R}^{D^h(L_{3d}+3L_{2d})}$&Feature obtained by interpolation of $\bm{x}^a$ in $\mathcal{H}$ \\
        $\bm{f}$ &$\mathbb{R}^{D}$&Any of anchor's attribute vectors $\in\{\bm{f}^a, \bm{l}, \bm{o}\}$ \\
        $\bm{\hat{f}}$ &$\mathbb{R}^{D}$&Quantized version of $\bm{f}$ \\
        $D$ &&Dimension of $\bm{f}$, which $\in\{D^a, 6, 3K\}$\\
        $\bm{q}$ &$\mathbb{R}^1$&Quantization step of $\bm{f}$ \\
        $\bm{r}$ &$\mathbb{R}^1$&Quantization step refinement term \\
        $\bm{\mu}$ &$\mathbb{R}^{D}$&Estimated mean value for distribution modeling \\
        $\bm{\sigma}$ &$\mathbb{R}^{D}$&Estimated standard deviation for distribution modeling \\
        $Q_0$ &&Base quantization step, which varies for $\bm{f}^a, \bm{l}, \bm{o}.$ \\
        $N$ &&Total number of anchors \\ 
        $h_f$ &&Occurrence frequency of ``+1'' in $\mathcal{H}$  \\ 
        $M_+$ &&Total number of ``+1'' $\mathcal{H}$ \\ 
        $M_-$ &&Total number of ``-1'' $\mathcal{H}$ \\ 
        $L_{\text{Scaffold}}$ &&The loss item used in Scaffold-GS~\cite{scaffold} \\
        $L_{\text{entropy}}$ &&Entropy loss for measuring bits of $\mathcal{A}$  \\
        $L_{\text{hash}}$ &&Entropy loss for measuring bits of $\mathcal{H}$  \\
        $L_m$ &&Masking loss  \\
        $Loss$ &&The total loss  \\
        $\lambda_e$ &&Tradeoff parameter to achieve variable birate \\
        $\lambda_m$ &&Tradeoff parameter to balance masking ratio \\
        ${\rm MLP_q}$ &&The MLP to deduce $r$ from $\bm{f}^h$ \\
        ${\rm MLP_c}$ &&The MLP to deduce $\mu$ and $\sigma$ from $\bm{f}^h$ \\
        $\phi$ &&Probability density function of Gaussian distribution  \\
        $\Phi$ &&Cumulative distribution function of Gaussian distribution  \\
    \bottomrule[2pt]
    \end{tabular}
    \label{tab:notation_table}
\end{table}

\clearpage

%
%
\end{document}